\documentclass[a4paper,conference]{IEEEtran}
\IEEEoverridecommandlockouts
\usepackage{times}  
\usepackage{helvet}  
\usepackage{courier}  
\usepackage[hyphens]{url}  
\usepackage{graphicx} 
\usepackage{algorithm}
\usepackage{algorithmic}
\usepackage{booktabs} 
\usepackage{adjustbox} 
\usepackage{amsmath}
\usepackage{comment}
\usepackage{array}
\usepackage{multirow}
\usepackage{xcolor}         
\ifCLASSINFOpdf
\else
\fi

\begin{document}
%
\title{DenseGAP: Graph-Structured Dense Correspondence Learning with Anchor Points}


\DeclareRobustCommand*{\IEEEauthorrefmark}[1]{%
  \raisebox{0pt}[0pt][0pt]{\textsuperscript{\footnotesize #1}}%
}

\author{%
  \IEEEauthorblockN{%
    Zhengfei Kuang\IEEEauthorrefmark{1,2},
    Jiaman Li\IEEEauthorrefmark{1,3},
    Mingming He\IEEEauthorrefmark{2}\textsuperscript{\textsection},
    Tong Wang\IEEEauthorrefmark{1,2} and
    Yajie Zhao\IEEEauthorrefmark{2}%
  }%
  \IEEEauthorblockA{%
  \IEEEauthorrefmark{1}University of Southern California
  \IEEEauthorrefmark{2}USC Institute for Creative Technologies \IEEEauthorrefmark{3} Stanford University}%
}


%


\maketitle

\begingroup\renewcommand\thefootnote{\textsection}
\footnotetext{Corresponding author.}

\begin{abstract}
Establishing dense correspondence between two images is a fundamental computer vision problem, which is typically tackled by matching local feature descriptors. However, without global awareness, such local features are often insufficient for disambiguating similar regions. And computing the pairwise feature correlation across images is both computation-expensive and memory-intensive. To make the local features aware of the global context and improve their matching accuracy, we introduce \textit{DenseGAP}, a new solution for efficient \textit{Dense} correspondence learning with a \textit{G}raph-structured neural network conditioned on \textit{A}nchor \textit{P}oints. Specifically, we first propose a graph structure that utilizes \textit{anchor points} to provide sparse but reliable prior on inter- and intra-image context and propagates them to all image points via directed edges. We also design a graph-structured network to broadcast multi-level contexts via light-weighted message-passing layers and generate high-resolution feature maps at low memory cost. Finally, based on the predicted feature maps, we introduce a coarse-to-fine framework for accurate correspondence prediction using cycle consistency. Our feature descriptors capture both local and global information, thus enabling a continuous feature field for querying arbitrary points at high resolution. Through comprehensive ablative experiments and evaluations on large-scale indoor and outdoor datasets, we demonstrate that our method advances the state-of-the-art of correspondence learning on most benchmarks. 
\end{abstract}


%
\IEEEpeerreviewmaketitle

\section{Introduction}
Image correspondence is the foundation of many computer vision tasks, such as geometric matching~\cite{DBLP:conf/wacv/MelekhovTSPRK19,DBLP:conf/cvpr/TruongDT20,DBLP:conf/nips/TruongDGT20}, pose estimation~\cite{sun2021loftr,DBLP:conf/cvpr/SarlinDMR20}, visual localization~\cite{DBLP:journals/corr/abs-2012-01909}, and optical flow~\cite{DBLP:conf/wacv/MelekhovTSPRK19,DBLP:conf/cvpr/TruongDT20,DBLP:conf/nips/TruongDGT20,DBLP:conf/iccv/JiangTHTY21}. Although being long explored, it remains an open question, especially for images under large appearance or view changes, or containing textureless or repetitive regions. The classic solution is based on keypoint detection and matching~\cite{DBLP:journals/ijcv/Lowe04,DBLP:journals/trob/LourencoBV12, DBLP:conf/cvpr/SarlinDMR20}. This line of methods is highly efficient but limited by the missing-detection issue~\cite{DBLP:conf/eccv/RoccoAS20}. Thus, the more recent works eliminate the dependency 
on keypoint detection by considering every point for building dense correspondence. 

Recent works on dense correspondence learning build 4D correlations between images using local features extracted for each point, followed by a neighbor consensus filtering strategy to select confident matches~\cite{DBLP:conf/nips/RoccoCATPS18,DBLP:conf/eccv/RoccoAS20,DBLP:conf/nips/LiHLP20,DBLP:conf/cvpr/LiHCHP20}. These methods are effective in finding denser matches but still suffer from two major limitations: (1) computing full points correlation is expensive and memory-intensive, especially on high-resolution images; (2) the extracted local features lack global context, making them indistinguishable in textureless or repetitive regions. The follow-up methods~\cite{DBLP:conf/nips/LiHLP20,DBLP:conf/eccv/WangZHS20} adopt coarse-to-fine frameworks to reduce the computational cost but struggle with the small receptive fields. To overcome them, we propose to utilize sparse correspondences as a bridge to connect every point in the global context, 
inspired by that humans typically use global information constructed by a few salient points to distinguish similar regions in a scene.

In this paper, we present a novel way of introducing sparse priors to dense correspondence learning using \textit{anchor points}, a set of paired salient points corresponded across images. With these anchor points, we propose to learn a context-aware feature field for querying correspondence at arbitrary image positions. We adopt a graph representation that connects the anchor points to every image position so as to model different levels of context and propagate them to the whole graph. Based on this representation, to integrate the global information into the local features, we further design three simple but effective message-passing layers: the inter-points layer binds anchor points to introduce the inter-image correlation, the intra-points layer aggregates information among anchor points within an image and builds the intra-image context, and the point-to-image layer broadcasts the above global contexts to every point and fuses it with the local features. Utilizing the predicted features, we finally present a coarse-to-fine framework to learn accurate dense correspondence based on cycle consistency. Extensive ablative experiments and comparisons show that our learned feature descriptors effectively boost the performance of dense correspondence prediction. In particular, our approach can help in complex tasks such as surface normal prediction, depth estimation, and object detection, where global context plays a critical role in extracting point-level information.

Our main contributions are summarised as follows: Firstly, we propose to use anchor points as priors for dense correspondence learning in a graph structure, which connects all local points in a global context. Secondly, we design a network based on the graph representation with three light-weighted message-passing layers for propagating and aggregating multi-level context information. Finally, our novel dense correspondence prediction pipeline achieves state-of-the-art (SOTA) performance, which supports arbitrary correspondence query for high-resolution input images and effectively embeds the global context to the local feature descriptors.

\begin{figure*}[h]
  \centering
  \includegraphics[width=0.85\textwidth]{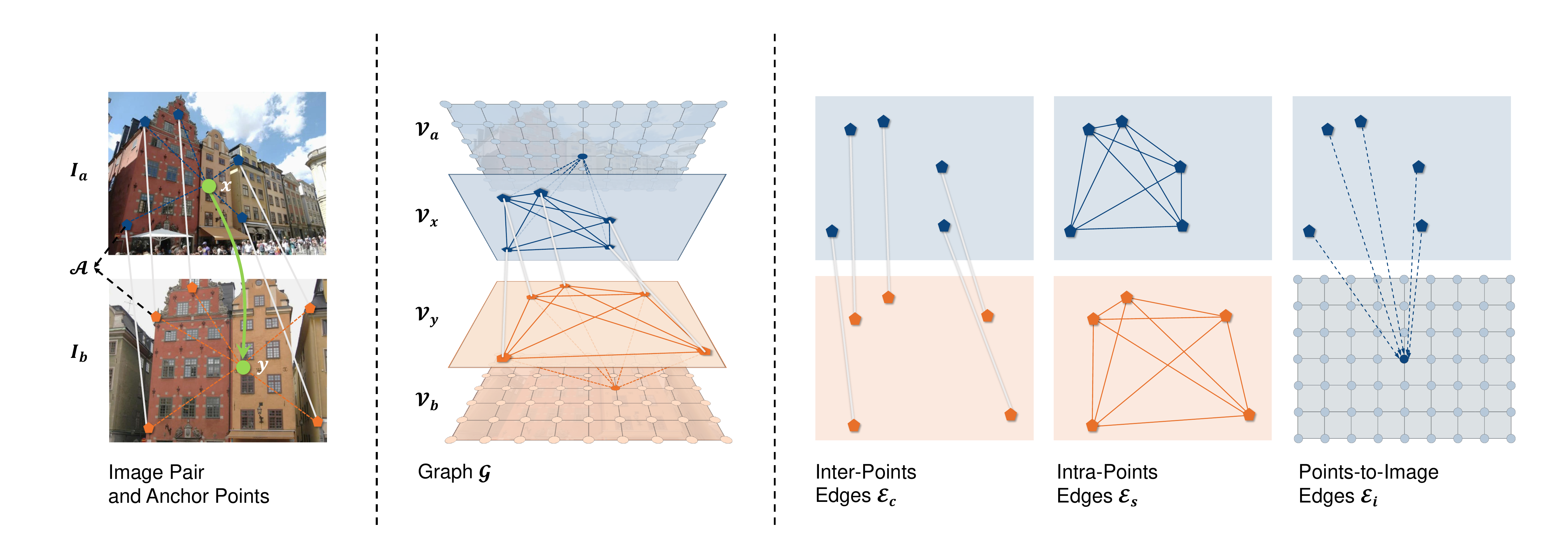}
  \vspace{-10pt}
  \caption{Illustration of the concept. From left to right: (1) Showcase of how anchor points guide our model to find dense correspondences; (2) Visualization of our designed graph; (3) Three types of edges in our graph. }
  \label{fig:concept}
  \vspace{-10pt}
\end{figure*}

\section{Related Work}

\paragraph*{Image Correspondence} The well-adopted pipeline for establishing image correspondence usually consists of feature detection~\cite{DBLP:journals/ijcv/Lowe04,DBLP:conf/iccv/LagunaRPM19,DBLP:conf/eccv/LencV16,DBLP:conf/eccv/MishkinRM18,DBLP:conf/cvpr/VerdieYFL15}, description~\cite{DBLP:conf/eccv/BayTG06,DBLP:conf/iccv/RubleeRKB11,DBLP:conf/iccv/LeuteneggerCS11,DBLP:conf/eccv/YiTLF16,DBLP:conf/nips/MishchukMRM17,DBLP:conf/cvpr/LuoSZZ0LFQ19,DBLP:conf/eccv/LuoSZZZYFQ18,DBLP:conf/nips/RevaudSHW19,DBLP:conf/cvpr/DusmanuRPPSTS19,DBLP:conf/cvpr/DeToneMR18}, and matching~\cite{DBLP:conf/cvpr/SarlinDMR20}. The typical drawback of these detector-based methods is the missing-detection problem, which limits the accuracy and the number of matches. To address this problem, detector-free approaches are explored. Some achieve feature matching by extracting features on a dense grid across the images~\cite{DBLP:conf/nips/RoccoCATPS18} and use coarse-to-fine frameworks to reduce memory footprint and improve fine-level matching ~\cite{DBLP:conf/eccv/RoccoAS20,DBLP:conf/nips/LiHLP20,DBLP:journals/corr/abs-2012-01909}. However, these frameworks require heavy computation of inter-image correlation and neglect the contextual cues. Another line of the detection-free methods~\cite{DBLP:conf/wacv/MelekhovTSPRK19,DBLP:conf/cvpr/TruongDT20,DBLP:conf/nips/TruongDGT20,DBLP:conf/cvpr/TruongDGT21} aims to generate pixel-level correspondence and bridge correspondence learning and optical flow estimation. They work well for continuous frames but are inadequate to handle image pairs with large displacements. 
Recently, the concurrent works~\cite{sun2021loftr, DBLP:conf/iccv/JiangTHTY21} involve global context between matches by using transformers~\cite{DBLP:conf/nips/VaswaniSPUJGKP17} which achieve great success in many NLP and vision tasks~\cite{DBLP:journals/corr/abs-2010-11929, DBLP:conf/eccv/CarionMSUKZ20, DBLP:conf/cvpr/YangYFLG20} using the attention mechanism. Different from them, we propose to adopt sparse correspondence as prior and design light-weighted network layers to efficiently propagate the contextual information to all image points, allowing predicting dense correspondence for arbitrary points.

\paragraph*{Graph-Structured Network} The graph-structured representation is applied in various domains, such as image~\cite{DBLP:conf/iccv/LiTLJUF17,DBLP:conf/cvpr/ChuangL0F18}, video~\cite{DBLP:conf/cvpr/VicolTCF18}, skeleton~\cite{DBLP:conf/aaai/YanXL18,DBLP:journals/tog/AbermanLLSCC20} and mesh~\cite{DBLP:conf/eccv/WangZLFLJ18,DBLP:conf/iccv/WenZLF19} thanks to the flexibility of this data structure. Meanwhile, more interests has put into relating graph representation with neural networks. The framework of graph neural network (GNN) is first proposed in~\cite{DBLP:journals/tnn/ScarselliGTHM09}, which formulates as node, edges, message-passing layers to assemble information from a graph structure. Inspired by GNN, some methods apply graph networks to vision tasks such as image recognition~\cite{DBLP:conf/cvpr/0004YG18}, object detection~\cite{DBLP:conf/cvpr/HuGZDW18}, point cloud learning~\cite{DBLP:journals/tog/WangSLSBS19} and so on. 
Our work introduces anchor points to bring the graph representation into correspondence learning. The graph representation is inspired by SuperGlue~\cite{DBLP:conf/cvpr/SarlinDMR20} which proposes a graph neural network for matching sparse keypoints between images. Different from SuperGlue~\cite{DBLP:conf/cvpr/SarlinDMR20}, we propose a more sophisticated graph to model multi-level contexts using sparse correspondence as prior and develop a general architecture to infuse the contextual information into local features. We follow the attention-based mechanism 
of Transformer~\cite{DBLP:conf/nips/VaswaniSPUJGKP17} to implement message-passing layers in the graph network, while Transformer~\cite{DBLP:conf/nips/VaswaniSPUJGKP17} is also used by recent works~\cite{sun2021loftr,DBLP:conf/iccv/JiangTHTY21} in a different way.

\section{Method}

\subsection{Anchor Points}
We propose to solve the problem of finding dense correspondence between a pair of images by first extracting a feature descriptor for arbitrary query points in one image and then using it to compute the correspondence in the other image. To efficiently encode the global information (\textit{e.g.} inter- and intra-image context) in the feature descriptor, we introduce anchor points to bridge all the points across images. The anchor points are a set of corresponding points from a pair of images that usually specify spatial locations of the salient features (\textit{e.g.} blobs, corners). They can be obtained by off-the-shelf sparse matching algorithms
(\textit{e.g.}~\cite{DBLP:conf/cvpr/SarlinDMR20,DBLP:conf/cvpr/LuoSZZ0LFQ19}), serving as reliable priors and modeling global contexts. Then we build a graph with anchor points and image points as nodes, connecting them with directed edges. By applying the message-passing mechanism~\cite{DBLP:journals/corr/abs-1806-01261}, we achieve the information propagation in the graph and aggregation for each node.

Given a pair of images $(I_a,I_b)$, and a normalized pixel coordinate $x \in [0,1]^2$ in $I_a$ as the query point, our target is to find its correspondence $y \in [0,1]^2$ in $I_b$. To achieve it, we adopt the approach introduced in \cite{DBLP:conf/eccv/WangZHS20} by extracting feature descriptors $F_a$ and $F_b$ of both images and computing $y$ as the expectation of the correlation-based distribution over $I_b$: 
\begin{equation}
\label{eq:softmax}
    y = \sum_{y \in I_b}y \cdot \text{Softmax}_y(F_a(x)^T F_b(y)).
\end{equation}
Note that $x\in[0,1]^2$ means a continuous coordinate and $x\in I_a$ indicates the pixel of $I_a$. To learn $F_a$ and $F_b$, we use anchor points, $\mathcal{A}_a\subset [0,1]^2$, $\mathcal{A}_b\subset [0,1]^2$ and their one-to-one correspondence $ {\mathcal{A}}=\{(x,y) | x\in\mathcal{A}_a, y\in\mathcal{A}_b\}$ as prior. We connect them with the image points in a directed graph ${\mathcal{G}}$, as shown in Fig.~\ref{fig:concept}. In $ {\mathcal{G}}$, we first build nodes for $ {\mathcal{A}}$, \textit{i.e.}, $ {\mathcal{V}}_x=\{v_x | x \in {\mathcal{A}_a}\}$ and $  {\mathcal{V}}_y=\{v_y | y \in {\mathcal{A}_b}\}$, and nodes for all image points of $I_a$ and $I_b$, \textit{i.e.}, ${\mathcal{V}}_a=\{v_x | x\in I_a\}$ and ${\mathcal{V}}_b=\{v_y | y\in I_b\}$. Then we connect them by three types of directed edges which we denote as $(v_s, v_r)$:
\begin{equation}
\begin{split}
     {\mathcal{E}}_{c} &= \{ (v_s, v_r) | (s, r) \in  {\mathcal{A}}\} \cup \{ (v_r, v_s) | (s, r) \in  {\mathcal{A}}\}, \\
     {\mathcal{E}}_{s} &= \{(v_s, v_r) | v_s,v_r \in  {\mathcal{V}}_x\} \cup \{(v_s, v_r) | v_s,v_r \in  {\mathcal{V}}_y\}, \\
     {\mathcal{E}}_{i} &= \{(v_s, v_r) | v_s \in  {\mathcal{V}}_x, v_r \in  {\mathcal{V}}_a\} \cup \\ & \quad \{(v_s, v_r) | v_s \in  {\mathcal{V}}_y, v_r \in  {\mathcal{V}}_b\}.
\end{split}
\end{equation}
${\mathcal{E}}_{c}$ indicate inter-points edges between anchor points from both images for inter-image communication; ${\mathcal{E}}_{s}$ represent intra-points edges between anchor points within the same image for intra-image communication; ${\mathcal{E}}_{i}$ are points-to-image edges from anchor points to image points, used to broadcast the information from anchor points to everywhere. Thus, the graph is represented as ${\mathcal{G}}=\left({\mathcal{V}}=\{{\mathcal{V}}_a, {\mathcal{V}}_b, {\mathcal{V}}_x, {\mathcal{V}}_y\}, {\mathcal{E}}=\{{\mathcal{E}}_{c}, {\mathcal{E}}_{s}, {\mathcal{E}}_{i}\}\right)$. 

We build a neural network based on this graph structure. Inspired by message-passing concept in graphical models, we design a message-passing layer for each type of edges ${\mathcal{E}}' \in {\mathcal{E}}$:
\begin{equation}
\label{eq:update_func}
\begin{split}
    z^{a}_r &= \sum_{(v_s, v_r)\in \mathcal{E'}}\alpha_{\mathcal{E}'}(z^{in}_s)\cdot \beta_{\mathcal{E}'}(z^{in}_s, z^{in}_r), \\
    z^{out}_r &= \rho_{\mathcal{E}'}(z^{in}_r, z^{a}_r).
\end{split}
\end{equation}
In detail, the message-passing layer first reprojects the input node attributes $z_s^{in}$ by $\alpha_{\mathcal{E}'}$, and then calculates the messages passed through the edges $(v_s, v_r)$ by $\beta_{\mathcal{E}'}$. Finally, it aggregates all information sent to the target node $v_r$ (denoted as $z_r^{a}$) and outputs the updated attributes by a feed-forward function ${\rho_{\mathcal{E}'}}$. All the functions in Eq.~\ref{eq:update_func} vary according to the edge types.

\subsection{Message-Passing Layers}
\label{sec:message_layer}
\paragraph*{Inter-Points Message-Passing Layer} 
This layer updates the features of anchor points using their counterparts in the other image through the edges ${\mathcal{E}}_{c}$. The anchor points are connected in a bipartite subgraph by edges ${\mathcal{E}}_{c}$. In this subgraph, all nodes have indegree and outdegree of $1$ as they are one-to-one paired. For this particular structure, we build a simple but effective layer by assigning the functions in Eq.~\ref{eq:update_func}:
\begin{equation}
\begin{split}
    \alpha_{\mathcal{E}_{c}}(z_s) &= 1, \\
    \beta_{\mathcal{E}_{c}}(z_s, z_r) &=  \mathcal{F}_{corr}(z_s \odot z_r), \\
    \rho_{\mathcal{E}_{c}}(z_r, z_{a}) &= z_r + z_{a},
\end{split}
\end{equation}
where $\odot$ means concatenation. The function first concatenates input features in a specified order (\textit{i.e.} $s$ first and then $r$), and then applies a two-layer multilayer perceptron (MLP) $\mathcal{F}_{corr}$ to get the aggregated information as the residual. The edges $(x_s, x_r)\in \mathcal{E}_{c}$ are existing in pairs, and thus this layer updates the features for anchor points in a symmetric way.

\paragraph*{Intra-Points Message-Passing Layer} This layer updates the feature descriptors of anchor points by aggregating messages across the edges ${\mathcal{E}}_{s}$. Each node is connected to all the others by ${\mathcal{E}}_{s}$ within the image, forming a complete subgraph. We update the node attribute based on the multi-head attention (MHA) used in~\cite{DBLP:conf/nips/VaswaniSPUJGKP17}, which has been proved a highly effective neural architecture in various mainstream vision tasks~\cite{DBLP:conf/eccv/CarionMSUKZ20,DBLP:journals/corr/abs-2010-11929}, including building sparse correspondence~\cite{DBLP:conf/cvpr/SarlinDMR20}. In our setting, each node is updated based on a weighted sum over its neighbours during the aggregation step (Fig.~\ref{fig:attention}). For each node $v_r$ connected by a set of incoming edges $(v_s, v_r)$, the first step is to generate the query vector $Q^h_r$ from $z_r$, and the key $K^h_s$ and value vector $V^h_s$ from $z_s$ for each head $h$. As in Eq.~\ref{eq:multi_head}, in each head, we sum up the value of all $z_s$ weighted by the attention $A^h_{s, r}$ calculated using the query and key vectors. Finally, we concatenate the aggregated value of all heads and use a feed-forward network $\mathcal{F}_{out}$ to refine the output. 
\begin{equation}
    \begin{split}
        Q^h_r &= \textbf{W}^h_q z_r, [K^h_s,V^h_s] = [\textbf{W}^h_k, \textbf{W}^h_v] z_s, \\
        A^h_{s, r} &= \text{Softmax}_s\left((K^h_s)^T Q^h_r / \sqrt{d_k}\right), \\
        V^h_r &= \sum_s A^h_{s, r}\cdot V^h_s, \\
        z_r^{out} &= \mathcal{F}_{out}(z_r+\textbf{W}_{out}(V^0_r\odot V^1_r\odot \ldots \odot V^h_r)),
    \end{split}
    \label{eq:multi_head}
\end{equation}
where $\textbf{W}_{q}, \textbf{W}_{k}, \textbf{W}_{v}, \textbf{W}_{out}$ are weight matrices, and $d_k$ is the dimension of $K^h_s$.

\begin{figure}[t]
    \centering
    \includegraphics[width=0.8\linewidth]{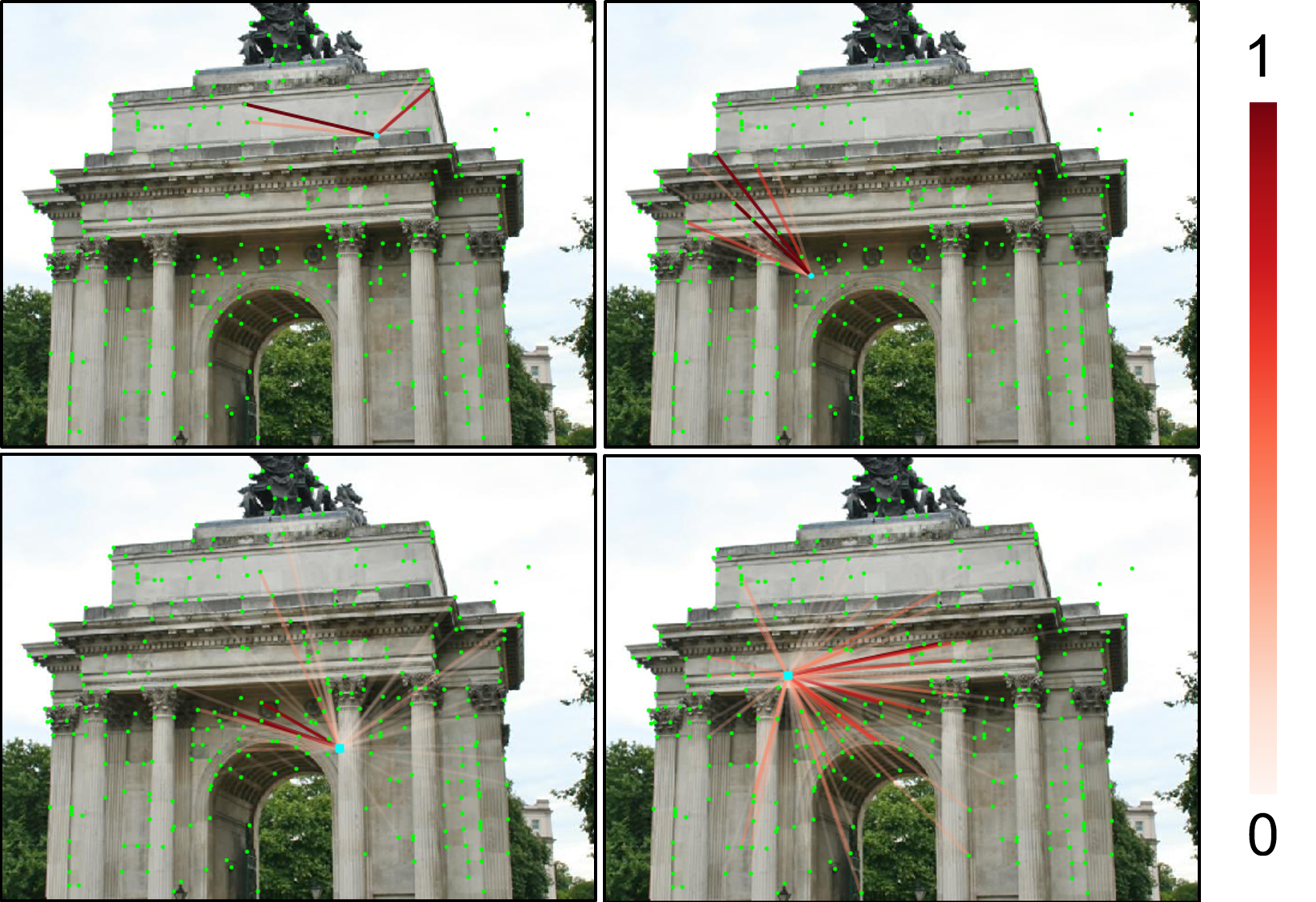}
    \vspace{-5pt}
    \caption{Visualization of attention. \textit{Row 1}: Attention in intra-points message-passing layer; \textit{Row 2}: Attention in point-to-image message-passing layer.}
    \label{fig:attention}
    \vspace{-15pt}
\end{figure} 

We adapt this attention model to our message-passing layer, where the functions in Eq.~\ref{eq:update_func} for this layer are defined as:
\begin{equation}
\label{Eq:E_s}
    \begin{split}
        \alpha^h_{\mathcal{E}_{s}}(z_s) &= V^h_s, \\
        \beta^h_{\mathcal{E}_{s}}(z_s, z_r) &= A^h_{s, r}, \\
        \rho_{\mathcal{E}_{s}}(z_r, z_{a}) &= \mathcal{F}_{out}(z_r+ \textbf{W}_{out}(z^0_{a}\odot z^1_{a}\odot \ldots \odot z^h_{a})). \\
    \end{split}
\end{equation}

\begin{figure*}[t]
  \centering
  \includegraphics[width=0.8\textwidth]{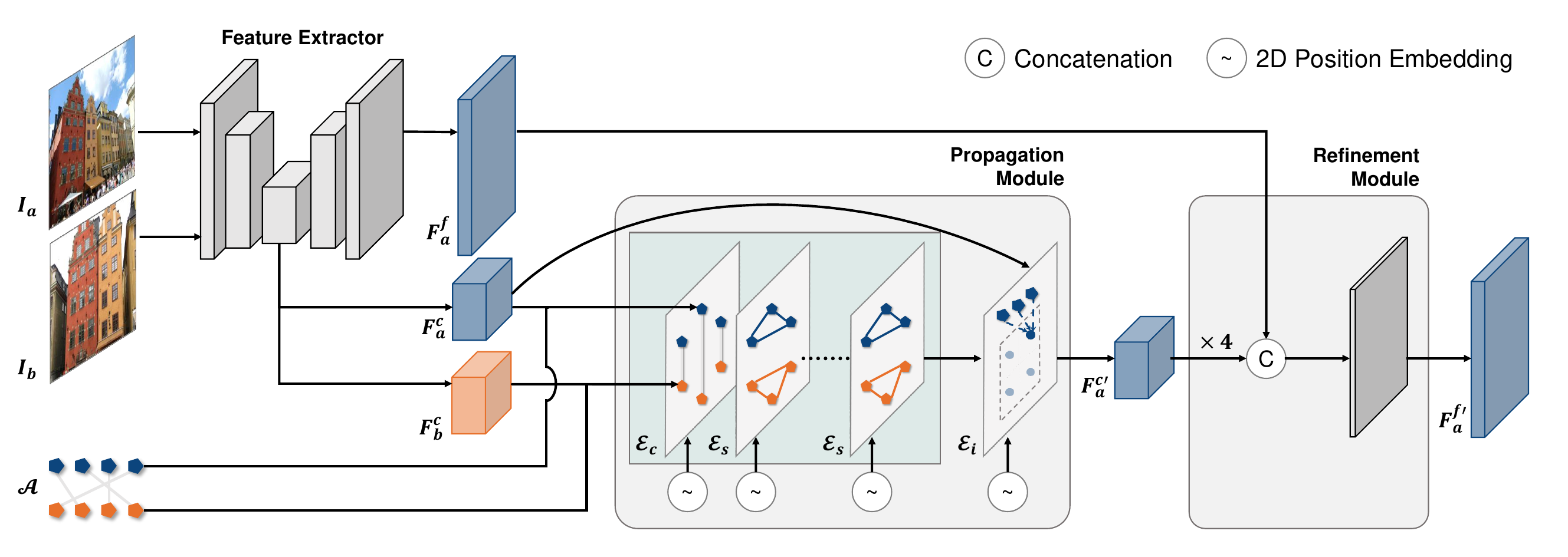}
  \vspace{-10pt}
  \caption{Overview of the framework. Given two images and anchor points, we first extract the coarse and fine feature maps of each image. Then we obtain the features of anchor points from the coarse feature maps as input to the Propagation Module. The output of the module is updated coarse feature maps, and is then fed with the fine feature maps to the Refinement Module. This module finally generates the updated fine feature maps.}
  \label{fig:overview}
  \vspace{-15pt}
\end{figure*}

\paragraph*{Points-to-Image Message-Passing Layer} This layer is designed to propagate the information learned by anchor points to all image points along the edges ${\mathcal{E}_{i}}$. Each pair of anchor point and image point are connected by only one directed edge in ${\mathcal{E}_{i}}$ and form a complete bipartite subgraph. The functions for ${\mathcal{E}_{i}}$ are the same as ones for ${\mathcal{E}_{s}}$ (Eq.~\ref{Eq:E_s}), while only the image points have their features updated in this layer by aggregating the updates from anchor points.

\subsection{Graph-Structured Network}
\label{sec:network}
Based on the graph structure, we design a network to learn feature descriptors conditioned on the input images ($I_a$ and $I_b$) and anchor points (${\mathcal{A}}_a$ and $\mathcal{A}_b$). The network contains two modules and updates the features in a coarse-to-fine manner. The propagation module integrates all message-passing layers and updates the features at the coarse level with larger receptive fields, which effectively reduces the computation cost while efficiently capturing global priors. The refinement module combines the updated coarse features and the local features at fine level to preserve the local structure details.

\begin{figure*}[t]
    \centering
    \begin{tabular}{ll}
        \begin{tabular}{l}
            \includegraphics[width=0.55\textwidth]{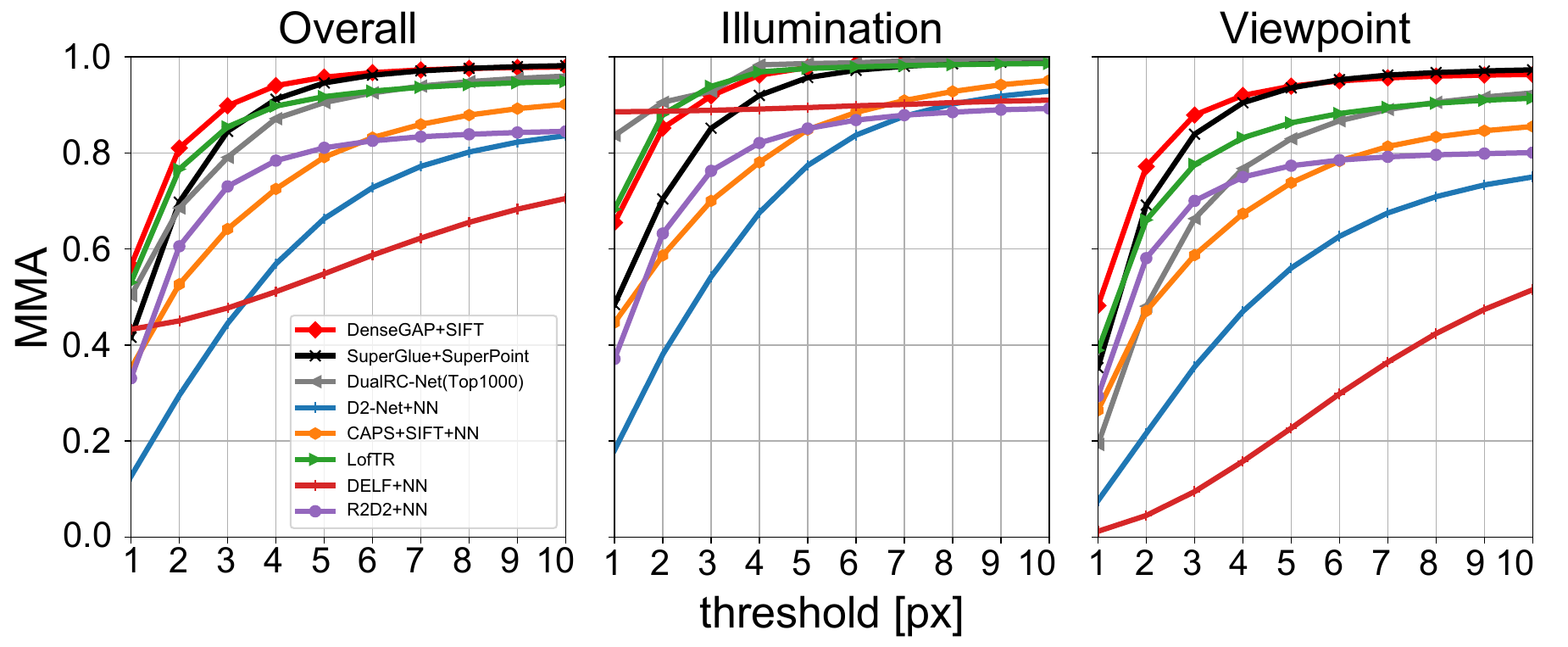} 
        \end{tabular} &
        \begin{adjustbox}{width=0.35\textwidth}
            \begin{tabular}{cc}
                \toprule
                Method  & \#Correspondences  \\
                \midrule
                D2-Net~\cite{DBLP:conf/cvpr/DusmanuRPPSTS19}+NN & 2.8K      \\
                DELF~\cite{DBLP:conf/iccv/NohASWH17}+NN & 1.9K     \\
                R2D2~\cite{DBLP:conf/nips/RevaudSHW19}+NN & 1.8K  \\
                CAPS~\cite{DBLP:conf/eccv/WangZHS20}+SIFT~\cite{DBLP:journals/ijcv/Lowe04}+NN & 1.5K  \\
                \midrule
                DualRC-Net~\cite{DBLP:conf/nips/LiHLP20}(Top1000) &  1.0K  \\
                SuperGlue~\cite{DBLP:conf/cvpr/SarlinDMR20}+SuperPoint~\cite{DeTone_2018_CVPR_Workshops} &  0.9K  \\
                LofTR~\cite{sun2021loftr} &  1.8K  \\
                DenseGAP+SIFT &  1.9K  \\
                \bottomrule
            \end{tabular}
        \end{adjustbox}
    \end{tabular}
  \vspace{-5pt}
  \caption{HPatches evaluation. Left: MMA comparison with previous work. Right: The mean number of correspondences for different methods.}
  \label{HPatches}
  \vspace{-10pt}
\end{figure*}

\begin{figure*}[t]
  \centering
  \begin{tabular}{ccc}
    \includegraphics[width=0.42\textwidth]{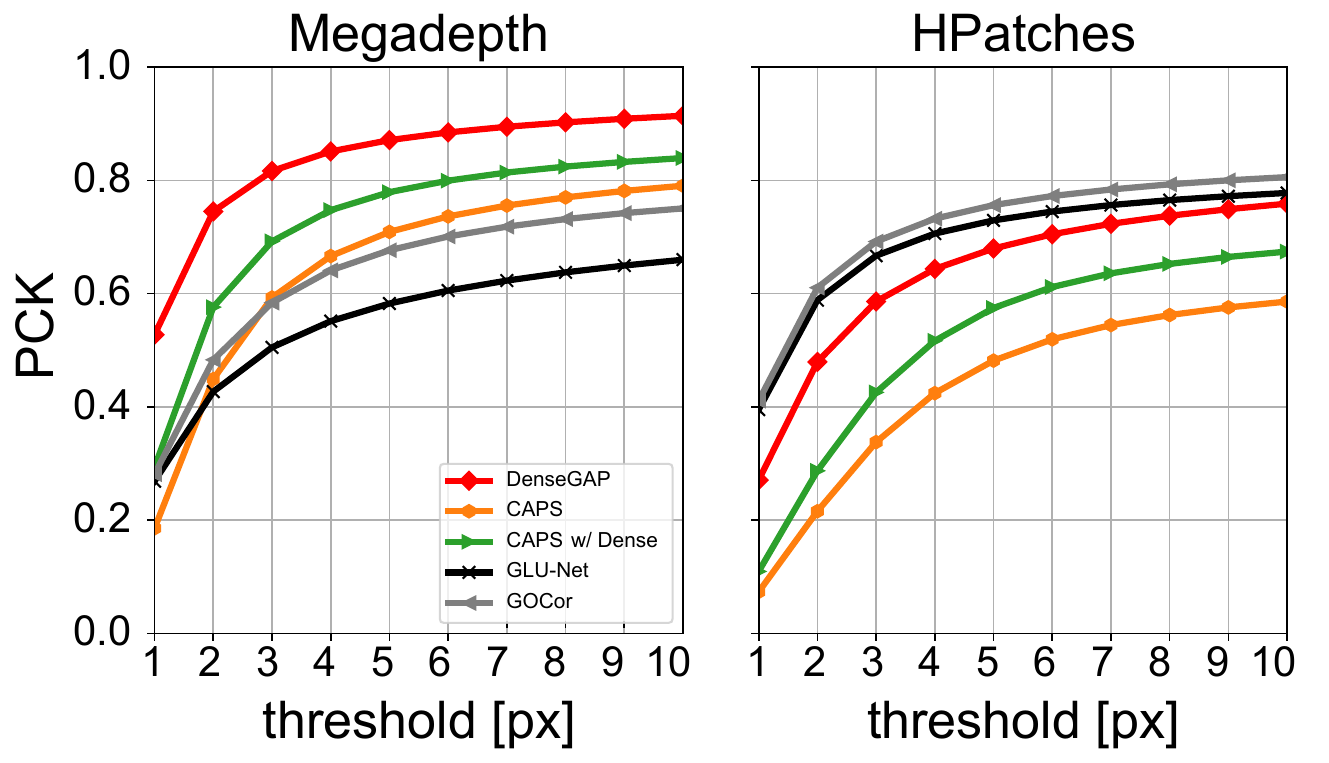} & 
    \includegraphics[width=0.3\textwidth]{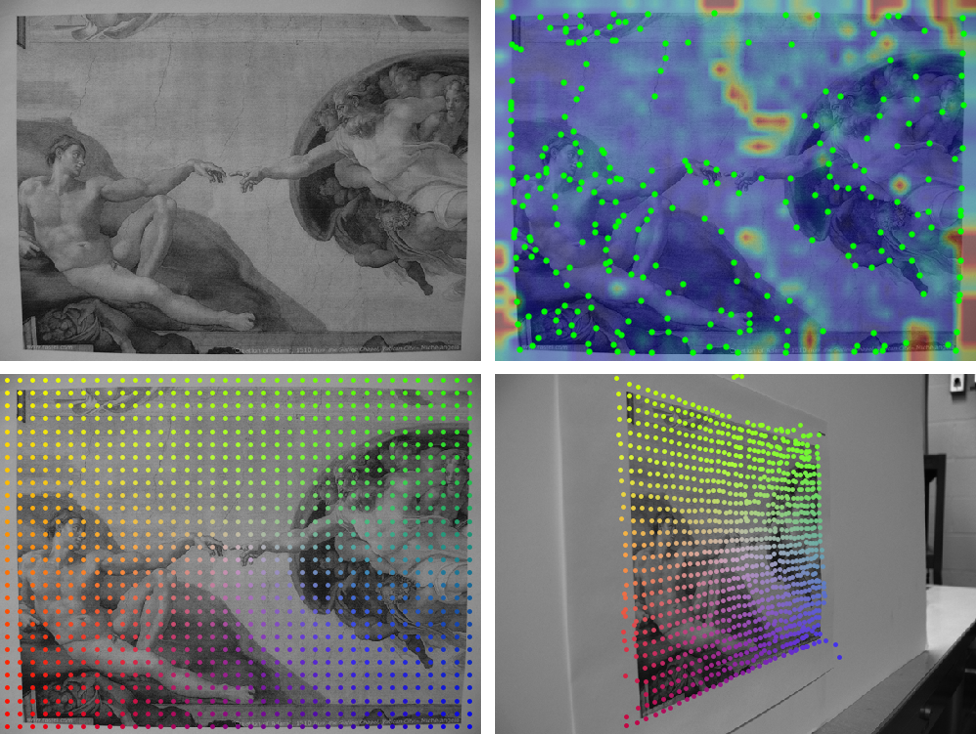} & 
    \includegraphics[width=0.05\textwidth]{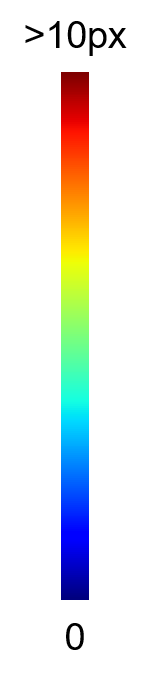} \\
    (a) PCK Evaluation & (b) Qualitative Results & 
  \end{tabular}
  \vspace{-2pt}
  \caption{Dense geometric matching evaluation. (a) Comparison of PCK scores. (b) Qualitative results on HPatches. \textit{Row 1}: \textit{Left}: Input Image, \textit{Right}: Error map of dense correspondences, and anchor points (colored in green) generated by SuperGlue~\cite{DBLP:conf/cvpr/SarlinDMR20}. We calculate the reprojection errors of each query point and generate the error map using bilinear interpolation. \textit{Row 2}: Correspondences between two images (indicated by different colors). The error bar on the right is only used for the error map. Note that we use the officially released pretrained model of GOCor~\cite{DBLP:conf/nips/TruongDGT20} and GLU-Net~\cite{DBLP:conf/cvpr/TruongDT20} in this experiment.} 
  \label{Dense-figure}
  \vspace{-10pt}
\end{figure*}

As shown in Fig.~\ref{fig:overview}, we first initialize local features at coarse and fine levels using a typical convolutional neural network (CNN), denoted as $F^c$ and $F^f$. Then we compute the features of anchor points by bilinearly interpolating $F^c$ and obtain the features $F^c_{{\mathcal{A}}_a}$ and $F^c_{\mathcal{A}_b}$ for anchor points in $I_a$ and $I_b$ respectively. Together with $F^c_a$ and $F^c_b$, they form the input of the propagation module, indicating the initial attributes of the nodes in ${\mathcal{V}}_x$, ${\mathcal{V}}_y$, ${\mathcal{V}}_a$, ${\mathcal{V}}_b$, which will be updated by the message-passing layers. The propagation module consists of $N_l$ intra-points message-passing layers and $N_l$ inter-points message-passing layers, followed by one point-to-image message-passing layer. We alternate the inter- and intra-points message-passing layers, starting with one inter-points message-passing layer. In addition, in each of the message-passing layers, we concatenate the node attributes with 2D position embeddings, which are calculated by the 2D sinusoidal position encoding method proposed in \cite{DBLP:conf/eccv/CarionMSUKZ20}. With the unique positional information, the learned features are position-dependent and more robust against matching ambiguity in indistinctive or textureless regions. This module finally outputs two updated coarse features ${F^c_a}'$ and ${F^c_b}'$, and feed them to the refinement module. In the refinement module, we bilinearly upsample ${F^c_a}'$ and ${F^c_b}'$ to fine level, concatenate them with the corresponding fine features and finally feed them to one convolutional layer to generate the result ${F^f_a}'$ and ${F^f_b}'$. 
\subsection{Coarse-to-Fine Training Strategy} 
\label{sec:training}

Since our network predicts both coarse- and fine-level feature maps, we use a coarse-to-fine matching strategy introduced by~\cite{DBLP:conf/eccv/WangZHS20} to compute the correspondence at a lower resolution followed by a local refinement at a finer scale. Given a query point $x$ in $I_a$, we first find its coarse correspondence $y_c$ using ${F^c_b}'$ and then crop a local window centered at $y_c$ in ${F^f_b}'$, extracting the final correspondence $y$ within the window.

\paragraph*{Losses} To train the model given an image pair, we randomly sample the query point $x$ from the pixels that can find ground-truth correspondences on the other image. For the set of training pairs $\mathcal{Q} = \{(x, y_{gt})\}$, the loss function is defined as the error between the established matches and ground-truth correspondence:
\begin{equation}
    \mathcal{L} = \sum_{(x, y_{gt})\in \mathcal{Q}} \frac{1}{\sigma_y}({\left\lVert y_{gt}-y_c \right\rVert}_2 + {\left\lVert y_{gt}-y \right\rVert}_2),
\end{equation}
where $\sigma_y$ is the uncertainty of the prediction proposed in~\cite{DBLP:conf/eccv/WangZHS20}. Besides, the anchor points are also randomly sampled from the points with known ground-truth correspondence. Meanwhile, we design a grid filter to make them evenly distributed. For more details, please refer to the supplementary material.

\paragraph*{Adaptive Position Embedding}
When training with fixed-size images, the learned model will degrade when testing with size-free images. To address this problem, we propose a simple and efficient method to augment the pixel coordinates with a random scale for each image. Specifically, in every training iteration, we assign a scale $r_a=(r_{a1}, r_{a2})$ for $I_a$, and $r_b=(r_{b1}, r_{b2})$ for $I_b$. Then, for every point $x=(x_1, x_2)$ from $I_a$, we scale it to $(x_1\cdot r_{a1}, x_2\cdot r_{a2})$ before feeding it to the position encoder, and 
apply to $I_b$ in the same way. This method significantly improves the result in size-free evaluations. 

\subsection{Runtime Correspondence Prediction}
\label{sec:testing}
At inference, we first extract anchor points for both input images, and then feed them to our graph-structured network to generate the feature maps. For any query point in $I_a$, we use the coarse-to-fine method same as training to compute its correspondence in $I_b$. Although the ground-truth correspondence anchor points are used for training, our model adapts well to the anchor points generated by other sparse matching methods when testing. In our experiments, we use SuperGlue~\cite{DBLP:conf/cvpr/SarlinDMR20} to efficiently provide reliable anchor points. Moreover, for any query point, we propose a metric based on cycle consistency to measure the confidence of its correspondence, which is used to filter the matches. The cycle consistency is defined as the euclidean distance between $x$ and $x'$ where $x$ is the query point in $I_a$ with its correspondence $y$ searched in $I_b$ and $x'$ is the matching point of $y$ when searching back in $I_a$.

\section{Experiments}
\label{experiments}

This section starts with training datasets and implementation details, followed by the evaluations of our approach on diverse tasks. Finally, we conduct a comprehensive ablation study of the proposed network structure. For training details and more results, please refer to the supplementary material.

\paragraph*{Datasets}
Our model is trained with MegaDepth~\cite{DBLP:conf/cvpr/LiS18} and ScanNet~\cite{DBLP:conf/cvpr/DaiCSHFN17} for outdoor and indoor scenes respectively.  MegaDepth~\cite{DBLP:conf/cvpr/LiS18} consists of over 600,000 preprocessed image pairs introduced by CAPS~\cite{DBLP:conf/eccv/WangZHS20}. We follow the same split of 130 scenes for training and 37 for validation. ScanNet is a large-scale indoor dataset, which is split into 1,513 training scenes and 100 testing scenes, same as~\cite{DBLP:conf/cvpr/SarlinDMR20}.  

\paragraph*{Implementation Details}
We adopt a modified ResNet-18~\cite{DBLP:conf/cvpr/HeZRS16} as backbone to extract feature maps. We set the number of layers $N_l$ to 4 and attention heads to 4 as well.  
When searching the correspondence in fine-level features, we set the window size as $1 /8$ of the feature map size.


\subsection{Image Matching}
\paragraph*{Datasets and Metrics}
HPatches is a benchmark dataset with 108 image sequences for evaluating the image matching accuracy. Each sequence contains one query image and five reference images with either changing illumination or viewpoints (52 for illumination and 56 for viewpoint).
We use mean matching accuracy (MMA)~\cite{DBLP:conf/cvpr/DusmanuRPPSTS19} for evaluating, which is calculated as the percentage of corrected matches in sampled query points within a threshold against ground truth matching. 

\paragraph*{Results}
We use keypoints extracted by SIFT~\cite{DBLP:journals/ijcv/Lowe04} as our query points, filter out all correspondences with cycle consistency larger than 5 pixels, and select top 2,000 matches for each image pair. We compare with R2D2~\cite{DBLP:conf/nips/RevaudSHW19}, D2-Net~\cite{DBLP:conf/cvpr/DusmanuRPPSTS19}, CAPS~\cite{DBLP:conf/eccv/WangZHS20}, SuperGlue~\cite{DBLP:conf/cvpr/SarlinDMR20}, LofTR~\cite{sun2021loftr} and DualRC-Net~\cite{DBLP:conf/nips/LiHLP20} and show that our model achieves the best overall performance with a large number of correspondences in Fig.~\ref{HPatches}. 

\subsection{Geometric Matching}
\label{sec:geometricmatching}
\paragraph*{Datasets and Metrics}
Both HPatches~\cite{DBLP:conf/cvpr/BalntasLVM17} (viewpoint sequences only) and MegaDepth~\cite{DBLP:conf/cvpr/LiS18} are used for this evaluation. We use the percentage of correct keypoints (PCK) as the evaluation metric. A correspondence is considered correct if it is close enough (\textit{e.g.} within a given threshold) to the ground truth. Following ~\cite{DBLP:conf/eccv/WangZHS20}, we densely sample correspondences between test image pairs and evaluate the PCK on them.   

\begin{table}
    \centering
    \caption{Pose Estimation Evaluation on the outdoor and indoor datasets. The * indicates the model trained on MegaDepth~\cite{DBLP:conf/cvpr/LiS18}.}
    \vspace{-5pt}
    \label{tab:pose_estimation}
    \begin{adjustbox}{width=\linewidth}
        \begin{tabular}{cccc}
            \toprule
            Method     & AUC(5) & AUC(10) & AUC(20)  \\
            \midrule 
            \multicolumn{4}{c}{MegaDepth~\cite{DBLP:conf/cvpr/LiS18}} \\
            \midrule
            DualRC-Net~\cite{DBLP:conf/nips/LiHLP20} & 32.56  & 47.60 & 61.40  \\
            SP~\cite{DeTone_2018_CVPR_Workshops}+SuperGlue~\cite{DBLP:conf/cvpr/SarlinDMR20} & 34.81 & 50.46  & 64.43       \\
            DenseGAP & \textbf{41.17} & \textbf{ 56.87} & \textbf{70.22}   \\            
            \midrule
            \multicolumn{4}{c}{ScanNet~\cite{DBLP:conf/cvpr/DaiCSHFN17}} \\
            \midrule
            DualRC-Net~\cite{DBLP:conf/nips/LiHLP20}* &  6.94 & 17.06 & 29.58 \\
            SP~\cite{DeTone_2018_CVPR_Workshops}+SuperGlue~\cite{DBLP:conf/cvpr/SarlinDMR20} & 16.16 & 33.81 & 51.84  \\
            DenseGAP*   & \textbf{16.93} & \textbf{34.85} & \textbf{53.16}  \\
            DenseGAP   & \textbf{17.01} & \textbf{36.07} & \textbf{55.66}  \\
            \bottomrule 
        \end{tabular}        
    \end{adjustbox}
    \vspace{-15pt}
\end{table}

\paragraph*{Results}

We show the results of our model compared to the SOTA methods (GOCor~\cite{DBLP:conf/nips/TruongDGT20}, GLU-Net~\cite{DBLP:conf/cvpr/TruongDT20} and CAPS~\cite{DBLP:conf/eccv/WangZHS20}) in Fig.~\ref{Dense-figure}. For a fair comparison, we also train CAPS ~\cite{DBLP:conf/eccv/WangZHS20} using our loss function, and test it separately (labeled as CAPS \textit{w/} Dense).
Our model (DenseGAP) significantly outperforms other methods on MegaDepth and achieves a comparable result on HPatches with GOCor~\cite{DBLP:conf/nips/TruongDGT20} and GLU-Net~\cite{DBLP:conf/cvpr/TruongDT20}. Both methods we think are naturally well-fit to predict displacements that can be interpolated bilinearly, such as the Homography space in HPatches. However, in a more general scenario with real-world non-planar objects (MegaDepth), DenseGAP outperforms by learning the distinctive features of each query point. 
Furthermore, while CAPS~\cite{DBLP:conf/eccv/WangZHS20} also uses a similar coarse-to-fine strategy to generate correspondences, DenseGAP achieves significant improvements thanks to the effectiveness of our graph-structured network. 

\subsection{Relative Pose Estimation}
\paragraph*{Datasets and Metrics}
We evaluate the model using pose estimation on MegaDepth~\cite{DBLP:conf/cvpr/LiS18} for outdoor scenes and ScanNet~\cite{DBLP:conf/cvpr/DaiCSHFN17} for indoor scenes. We randomly select 2,459 image pairs from the validation dataset of MegaDepth and use 1,500 image pairs of ScanNet~\cite{DBLP:conf/cvpr/DaiCSHFN17} provided by~\cite{DBLP:conf/cvpr/SarlinDMR20}.
We adopt the same evaluation metric as~\cite{DBLP:conf/cvpr/SarlinDMR20}, which calculates the area under cumulative error curve (AUC) of pose error up to thresholds ($5^\circ$, $10^\circ$, $20^\circ$). The relative pose is estimated by applying RANSAC~\cite{DBLP:journals/cacm/FischlerB81} on the correspondences. 

\paragraph*{Results}
We compare with the SOTA dense correspondence method DualRC-Net~\cite{DBLP:conf/nips/LiHLP20} and the SOTA sparse matching method SuperGlue~\cite{DBLP:conf/cvpr/SarlinDMR20} in Tab.~\ref{tab:pose_estimation}. We use 500 matching pairs of the SuperGlue output as the anchor points, and query the SIFT~\cite{DBLP:journals/ijcv/Lowe04} keypoints and dense points following the same sampling strategy as in Sec~\ref{sec:geometricmatching}. We filter the matches by selecting the top 8,000 correspondences from the predicted correspondences with cycle consistency larger than 5 pixels. Compared to SuperGlue, our model significantly boosts the performance using denser matches. We attribute this improvement to successfully getting dense correspondences based on sparse priors, which allows us to extract pixel-level image information and fuse them with contextual information from anchor points, thus leading to less biased results.


    
    
    
    

\subsection{Ablation Study}

\begin{figure}[t]
    \centering
    \includegraphics[width=0.9\linewidth]{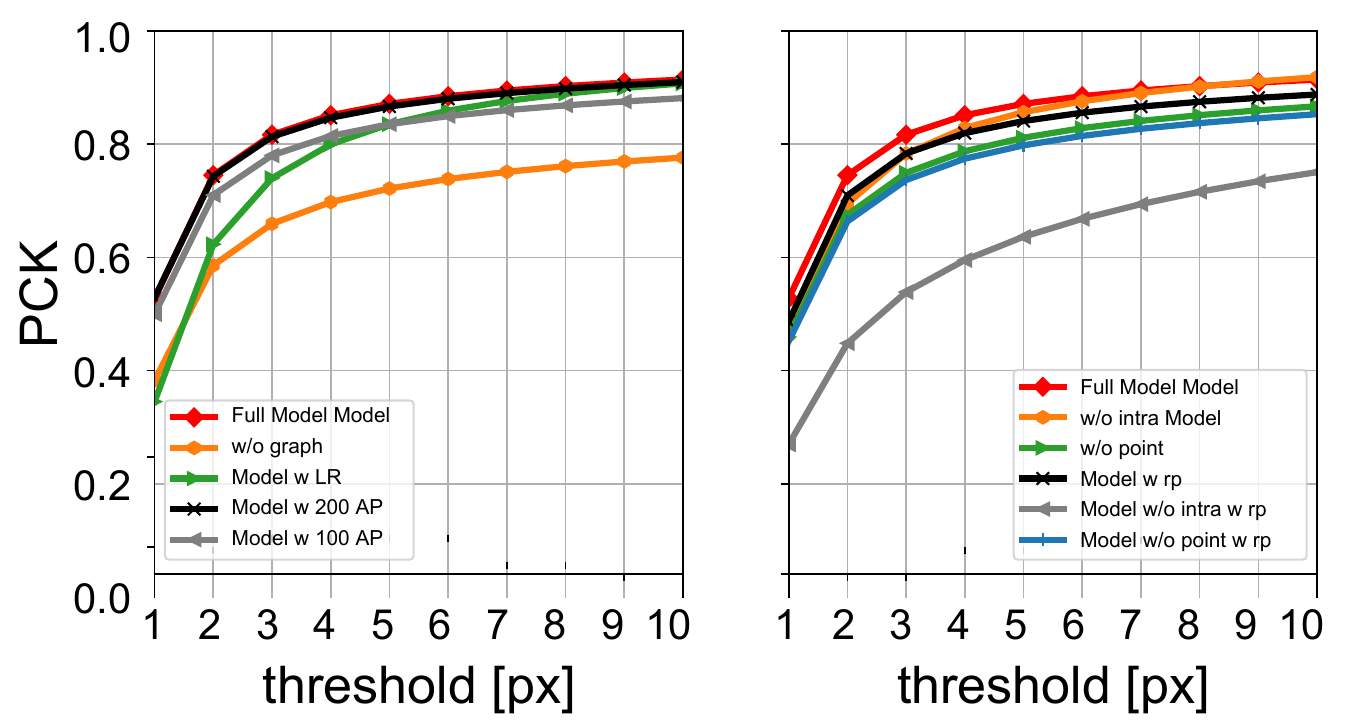}
    \vspace{-10pt}
    \caption{Our model in different settings (left) and using different combinations of message-passing layers (right).}
    \label{fig:ablation}
    \vspace{-15pt}
\end{figure}

We conduct two ablation studies on MegaDepth with the PCK metric in Fig.~\ref{fig:ablation}.
We first study the performance of the proposed model under different settings (left). It consists of four variants:
(1) \textit{Model w/o graph} removes the graph-structured representations and only preserves the local feature extractor; (2) \textit{Model w/ lower resolution (LR)} changes the resolution of feature maps to $1/4$ and $1/16$ of the image size; and (3)\&(4) \textit{Models w/ 200/100 anchor points(AP)} reduce the number of anchor points to 200 and 100, to show the performance of limited anchor points. The first two variants decrease the score in different patterns, which indicates the feasibility and inevitability of our design. Reducing the number of anchor points does not have much effect on the results unless the number is too low, indicating that our model is robust to the number of anchor points. Then we explore the effectiveness of our message-passing layers using two variants of our model under the original setting (right in Fig.~\ref{fig:ablation}): (1) \textit{Model w/o intra} removes the intra-points layer; and (2) \textit{Model w/o point} removes both intra-points and inter-points layers. Additionally, we test them with a more challenging setting where the locations of $60\%$ of anchor points are interfered 
with Gaussian noise with standard deviation of 50 pixels (\textit{i.e.}, \textit{Model w/o intra w/ rp}, \textit{Model w/o point w/ rp}, \textit{Full Model w/ rp}). We observe that without the intra-points layer, the model performance is close to the full model in the original setting, but substantially degrades as the outliers increase. The model without inter- and intra-points layers performs obviously worse than the full model due to the lack of cross-image context.

\section{Conclusion}
\label{conclusion}
We propose a novel dense correspondence learning approach that utilizes anchor points with a graph-structured network. The feature descriptors fusing contextual information introduced by anchor points with local information serve for correspondence establishment for any query point and significantly improve the performance on diverse tasks. This model has the potential to generalize to other tasks such as normal estimation, optical flow, etc. An end-to-end solution will be an interesting future direction that jointly optimizes anchor points and dense correspondence.


\section*{Acknowledgements}
This research is sponsored by the Department of the Navy, Office of Naval Research under contract number N00014-21-S-SN03 and in part sponsored by the U.S. Army Research Laboratory (ARL) under contract number W911NF-14-D-0005. Army Research Office sponsored this research under Cooperative Agreement Number W911NF-20-2-0053. Statements and opinions expressed, and content included, do not necessarily reflect the position or the policy of the Government, and no official endorsement should be inferred. The views and conclusions contained in this document are those of the authors and should not be interpreted as representing the official policies, either expressed or implied, of the Army Research Office or the U.S. Government. The U.S. Government is authorized to reproduce and distribute reprints for Government purposes notwithstanding any copyright notation herein.

\bibliographystyle{unsrt}
\fontsize{9.0pt}{10.0pt} \selectfont
\bibliography{arXiv}

\end{document}